\documentclass[a4paper, 10pt, conference]{IEEEtran}

\IEEEoverridecommandlockouts

\usepackage{amsmath, amssymb}
\usepackage[pdftex]{graphicx}
\graphicspath{{figures/}}
\usepackage{hyperref}
\hypersetup{
  colorlinks=true,
  linkcolor=black,
  citecolor=black,
  filecolor=black,
  urlcolor=black,
}
\usepackage[T1]{fontenc}
\usepackage[utf8]{inputenc}
\usepackage{csquotes}
\usepackage[english]{babel}
\usepackage[font=small]{caption, subcaption}
\usepackage{siunitx}
\usepackage{booktabs, environ, pgfplots, pgfplotstable, tikz}

\makeatletter
\newsavebox{\measure@tikzpicture}
\NewEnviron{scaletikzpicturetowidth}[1]{
  \def\tikz@width{#1}
  \def\tikzscale{1}\begin{lrbox}{\measure@tikzpicture}
  \BODY
  \end{lrbox}
  \pgfmathparse{#1/\wd\measure@tikzpicture}
  \edef\tikzscale{\pgfmathresult}
  \BODY
}
\makeatother

\title{
  Single-Stage Object Detection from Top-View Grid Maps on Custom Sensor Setups
}

\author{
\IEEEauthorblockN{Sascha Wirges\IEEEauthorrefmark{1} \& Shuxiao Ding}
\IEEEauthorblockA{
Mobile Perception Systems Group\\
FZI Research Center for Information Technology\\
Karlsruhe, Germany\\
\{wirges, ding\}@fzi.de
}
\and
\IEEEauthorblockN{Christoph Stiller}
\IEEEauthorblockA{
Institute of Measurement and Control Systems\\
Karlsruhe Institute of Technology (KIT)\\
Karlsruhe, Germany\\
stiller@kit.edu}
}

\begin{document}

\maketitle
\thispagestyle{empty}
\pagestyle{empty}

\begin{abstract}
We present our approach to unsupervised domain adaptation for single-stage object detectors on top-view grid maps in automated driving scenarios.
Our goal is to train a robust object detector on grid maps generated from custom sensor data and setups.
We first introduce a single-stage object detector for grid maps based on RetinaNet.
We then extend our model by image- and instance-level domain classifiers at different feature pyramid levels which are trained in an adversarial manner.
This allows us to train robust object detectors for unlabeled domains.
We evaluate our approach quantitatively on the nuScenes and KITTI benchmarks and present qualitative domain adaptation results for unlabeled measurements recorded by our experimental vehicle.
Our results demonstrate that object detection accuracy for unlabeled domains can be improved by applying our domain adaptation strategy.
\end{abstract}

\section{Introduction} \label{sec:introduction}

A detailed environment model is an essential part of scene understanding modules of automated driving systems.
These systems require a robust and reliable detection of other traffic participants such as cars, cyclists and pedestrians as subsequent tasks such as motion planning rely on this information.
Here, \textit{object detection} refers to the detection, shape estimation and semantic classification of relevant objects.
In this work, we represent the environment by top-view grid maps.
Their structure enables the fusion of heterogeneous range sensor data (e.g. lidar and radar) and real-time capability at bounded memory and computation.

\begin{figure}[ht]
    \centering
    \def\svgwidth{\columnwidth}
    \begin{footnotesize}
        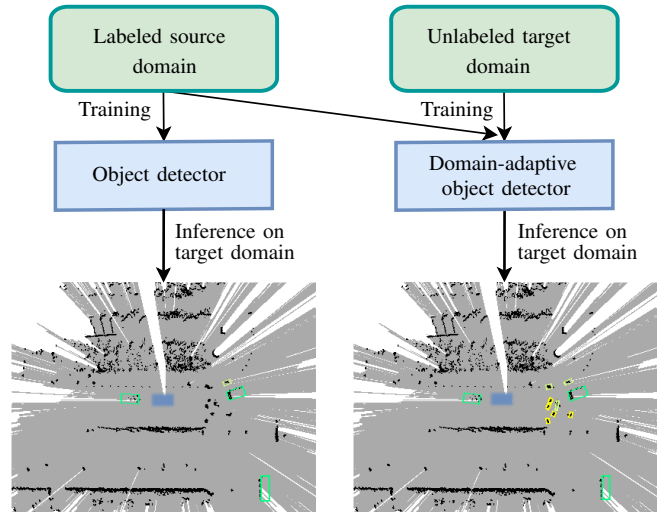
    \end{footnotesize}
    \caption{
        Qualitative comparison of cross-domain object detection trained without and with our adaptation strategy.
        During training, we align domain-dependent feature distributions by considering examples from a labeled source and an unlabeled target domain.
        We observe that our domain-adaptive object detector achieves a higher accuracy and a much higher recall for target domain examples.
        Cars, cyclists and pedestrians are depicted in green, yellow-green and yellow, respectively.
        The ego vehicle (blue) is depicted in the center. 
    }
    \label{fig:overview}
\end{figure}

In recent years, many convolutional object detectors applied to the automated driving domain were proposed that increased the accuracy on benchmarks such as KITTI \cite{Geiger2012}, Cityscapes \cite{Cordts2016} and nuScenes \cite{Caesar2019}.
Here, we are interested in maximizing the object detection accuracy for our own sensor setup.
However, due to the domain gap between measurements recorded with a custom sensor setup and publicly available data sets the object detection performance drops significantly.

Acquiring a data set that fits all testing scenarios is impractical due to high annotation expenditure.
Therefore, learning a model from labeled source domain examples and applying it to examples of another domain, referred to as \textit{domain adaptation}, is necessary to accomplish accurate object detection on custom data.
The domain discrepancy in top-view grid maps is mostly caused by different sensor types or setups.
Traffic situation and city layout where data was recorded is also an important factor.

In this paper, we propose an approach of domain-adaptive object detection for top-view multi-layer grid maps which mitigates a performance deterioration due to the domain gap caused by custom sensor setups.
We first present a real-time capable single-stage object detector for grid maps based on RetinaNet \cite{Lin2017a}.
We then extend the detector with image- and instance-level domain classifiers at different feature pyramid levels and apply an adversarial loss which enforces intermediate feature representations to be domain-invariant.
Our hypothesis is that the resulting object detector shows improved performance on target domain examples.

After presenting recent work on object detectors and domain adaptation strategies we compare two different benchmarks used in this paper towards their domain gap in Sec.~\ref{sec:related_work}.
We then present our object detector and highlight structural changes and principles in order to train it in an adversarial manner for domain adaptation (Sec.~\ref{sec:object_detection_model}).
The training details including loss function and hyperparameters are presented in Sec.~\ref{sec:training}.
We provide quantitative evaluation results and an ablation study in Sec.~\ref{sec:evaluation} and show qualitative results when our object detector is applied to data recorded from our own experimental vehicle.
Finally, we conclude our findings and highlight future work in Sec.~\ref{sec:conclusion}.

\section{Related Work} \label{sec:related_work}

\subsection{Object Detection} \label{sec:related_work_object_detection}
In the last years many object detection models have been proposed which can be divided into two families, two-stage detectors such as Faster-RCNN \cite{Ren2017} and single-stage detectors such as SSD \cite{Liu2016}.
Single-stage detectors do not need an additional region-proposal network and are usually faster while still achieving sufficiently good performance.
RetinaNet \cite{Lin2017a} uses a Feature Pyramid Network (FPN) \cite{Lin2017b} as feature extractor which combines different pyramid levels by combining top-down and bottom-up feature extraction, significantly improving the detection of small objects.

Object detection research in automated driving focuses mostly on 3D or top-view detection of other traffic participants.
The top-view representation projects 3D measurements onto a common ground surface, hence 2D convolutions can be applied which improve the computation efficiency.
For instance, TopNet \cite{Wirges2018}, Complex-YOLO \cite{Simon2018} and BirdNet \cite{Beltran2018} preprocess range measurements by projecting them onto a top-view ground plane.
PIXOR \cite{Yang2018a} and HDNET \cite{Yang2018b} compute occupancy features at different heights from lidar point sets which minimize the information loss due to ground surface projection.
PointPillars \cite{Lang2019} proposes a fast encoding of low-level point set features along pillars to create a top-view feature representation.
In this work, we use grid maps as environment representation which are generated using fixed hand-crafted features, however, we stress that our method can also be applied to structurally different models.

\subsection{Domain Adaptation} \label{sec:related_work_domain adaptation}
Many domain adaptation approaches are based on generative adversarial networks (GANs) \cite{Goodfellow2014}, e.g. image-to-image translation \cite{Isola2017} and CycleGAN \cite{Zhu2017}.
A GAN consists of a feature generator, a discriminator to distinguish these features and an adversarial loss which enforces the generator to produce indistinguishable features.
Research on domain-adaptive object detection mostly focuses on aligning the intermediate feature distributions using adversarial methods.
Based on the fundamentals introduced in \cite{Ben-David2010}, \cite{Ganin2015} measures the $\mathcal{H}\Delta\mathcal{H}$-divergence as domain discrepancy using a classifier.
By flipping the sign of gradients from the domain classifier branch, the feature extractor is trained in an adversarial manner to maximize the domain classification loss.
This method reduces the domain discrepancy, resulting in domain-invariant features.
Ganin et al. efficiently implemented their approach in a gradient reversal layer(GRL).

Chen et al. \cite{Chen2018} employ domain classifiers on image- and instance-level features in Faster R-CNN as well as a consistency regularization.
Other approaches for unsupervised domain-adaptive object detection are also built on Faster R-CNN, e.g. Strong-Weak \cite{Saito2019} and Stacked Complementary Losses (SCL) \cite{Shen2019}.

\subsection{Benchmarks and Data Sets} \label{sec:related_work_datasets}
\subsubsection{KITTI Bird's Eye View Benchmark}
The KITTI Bird's Eye View Evaluation 2017 \cite{Geiger2012} is a benchmark for object detection from top-view images which consists of 7481 train and 7518 test examples collected in southwest Germany.
In addition to camera images, lidar measurements and sensor calibration are available.
Train and test set include 80,256 labeled objects which are represented by one of eight semantic classes and seven 3D bounding box parameters.
The benchmark evaluates the object classes \textit{car}, \textit{pedestrian} and \textit{cyclist} separately.
During evaluation, the average precision at 40 recall positions is computed and referred to as \textit{average precision} (AP).
Object detections are considered positive matches if the Intersection-over-union (IoU) threshold is larger than 70\% for cars and 50\% for cyclists and pedestrians.

\subsubsection{nuScenes}
The nuScenes data set \cite{Caesar2019} consists of 1000 driving scenes with a duration of 20 seconds, collected in Boston and Singapore.
The data set includes approximately 1.4M camera images, 390k LIDAR sweeps, 1.4M RADAR sweeps and 1.4M labeled 3D bounding boxes from 40k key frames.
The 3D object detection benchmark considers ten different object classes, e.g. \textit{truck}, \textit{car}, \textit{motorcycle}, \textit{bicycle} or \textit{pedestrian}.

In contrast to KITTI, nuScenes uses the 2D center distance in the top-view plane instead of the box IoU to define a box match.
Object classifications can be evaluated class-independent by calculating the AP and true positive metrics such as the \textit{average translation error} (ASE) or the \textit{average scale error} (ASE).
The \textit{mean average precision} (mAP) is calculated by averaging over the AP within a distance threshold of \{0.5, 1, 2, 4\} meters.

\subsubsection{Comparison}
Tab.~\ref{tab:dataset} briefly summarizes the differences between the KITTI and nuScenes data set.
In comparison to KITTI, nuScenes has 7 times more annotations.
We also observe a domain gap in the distribution of different object classes and in the resulting grid maps.

\begin{table}[ht]
    \centering
    \begin{tabular}{l|c|c}
        \textbf{Data set}               & \textbf{KITTI} \cite{Geiger2012} & \textbf{nuScenes} \cite{Caesar2019} \\
        \hline
        \textbf{City}                   & Southwest Germany                & Singapore, Boston                   \\
        \textbf{Range sensor}           & Velodyne HDL64E-S2               & Unspec. 32 line lidar               \\
        \textbf{Eval. categories}       & 3                                & 10                                  \\
        \textbf{Training set size}      & 7481                             & 28,130                              \\
        \textbf{Cars}                   & 28,742 (70.8\%)                  & 493,322 (42.3\%)                    \\
        \textbf{Pedestrians}            & 4487 (11.1\%)                    & 222,164 (19.1\%)                    \\
        \textbf{Cyclists}               & 1627 (4.0\%)                     & 11,859 (2.1\%)                      \\
    \end{tabular}
    \caption{
        Comparison between KITTI object detection and nuScenes data set.
        We include only the training set quantities of cars, pedestrians and cyclists.
    } \label{tab:dataset}
\end{table}

\section{Object Detection Model} \label{sec:object_detection_model}

\begin{figure*}[ht]
    \centering
    \def\svgwidth{1.95\columnwidth}
    \begin{footnotesize}
        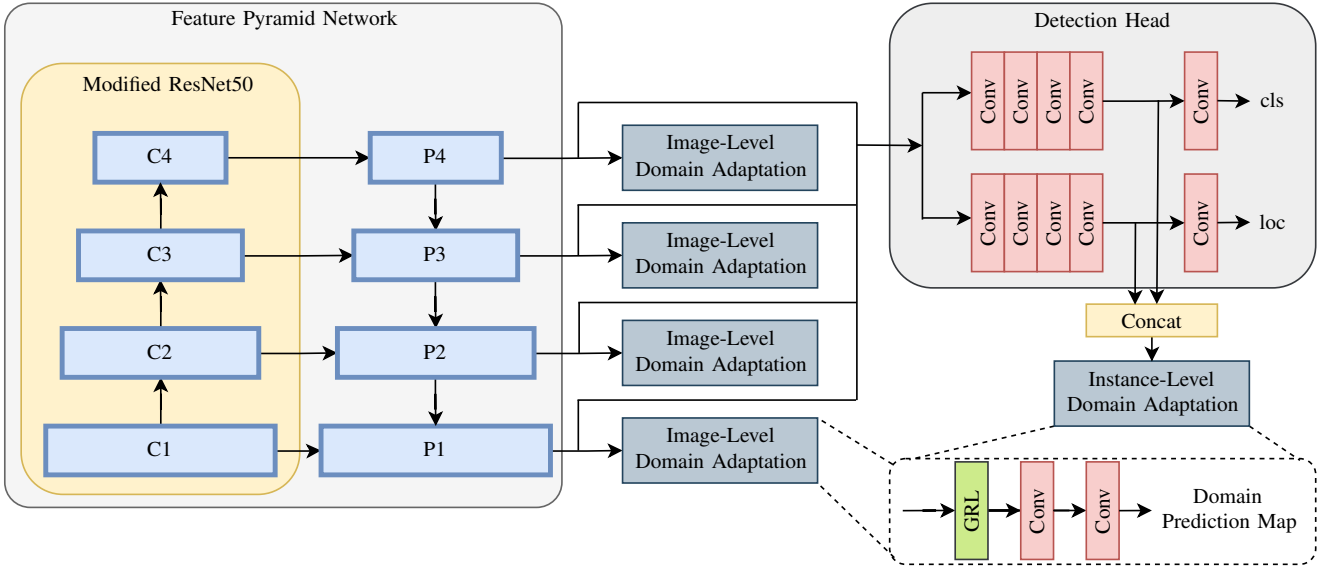
    \end{footnotesize}
    \caption{
        System overview of our single stage object detector with hierarchical image- and instance-level domain adaptation, following \cite{Chen2018}.
        The image-level domain classifiers are attached to FPN feature maps at different pyramid levels.
        The instance-level domain classifier belongs to the detection header.
        All classifiers have identical structure.
    } \label{fig:network}
\end{figure*}

A system overview of our domain-adaptive object detector is depicted in Fig.~\ref{fig:network}.
The entire model can be divided into feature extractor, detection head and domain classifier.
The first two parts correspond to a single-stage object detector whereas the adversarial domain classifier is only applied during training.

\subsection{Single Stage Detector} \label{sec:single_stage_detector}
Our object detector follows RetinaNet \cite{Lin2017a} with modifications designed for top-view grid maps as input.
Following previous work \cite{Wirges2018, Wirges2019}, we use five grid cell features:
\begin{itemize}
    \item Number of reflections
    \item Height difference of reflections
    \item Average reflected energy
    \item Number of transmissions (via ray-casting) 
    \item Height of occlusions (via ray-casting)
\end{itemize}
Throughout this work we use a grid cell size of 15cm and a total range of 60m$\times$60m.

We process the different grid cell features separately by applying depth-wise separable convolutions in the first layer.
As feature extractor we adapt a ResNet50 \cite{He2016} with reduced filter sizes.
Small objects in grid maps complicate the target assignment between anchors and ground truth boxes, leading to bad classification performance.
To mitigate this problem, we reduce the amount of downsampling operations by using only the pyramid feature maps $\boldsymbol{P}_1, \boldsymbol{P}_2, \boldsymbol{P}_3, \boldsymbol{P}_4$.
A shared-weight detection head handles both classification and box regression tasks on all pyramid levels.
Following the encoding discussed in \cite{Wirges2018}, we encode a rotated bounding box $x_\mathrm{c}, y_\mathrm{c}, w, h, \theta$ by the six parameters
\begin{equation*}
    \frac{x_\mathrm{c}-x_\mathrm{a}}{w_\mathrm{a}}, \frac{y_\mathrm{c}-y_\mathrm{a}}{h_\mathrm{a}}, \log \left( \frac{w}{w_\mathrm{a}} \right), \log \left( \frac{h}{h_\mathrm{a}} \right), \sin \left( 2\theta \right), \cos \left( 2\theta \right)
\end{equation*}
with respect to the anchor parameters $x_\mathrm{a}, y_\mathrm{a}, w_\mathrm{a}, h_\mathrm{a}$.
Based on the average object size in the real world \cite{Wirges2018}, we define the basis anchor sizes $8^2, 16^2, 32^2, 64^2$ on different pyramid levels.
We use the three aspect ratios $(1:2), (1:1), (2:1)$ and two anchor scales $2^0, 2^{1/2}$ at each pyramid level.

\subsection{Domain Adaptation Strategy} \label{sec:domain_adaptation_strategy}
We aim to extract domain-invariant features that are still well-suited for object detection.
Therefore, we adapt the approach of Chen et al. \cite{Chen2018} who distinguish between image- and instance-level domain adaptation.
Our domain classifiers consist of two convolutional layers, producing one output followed by a sigmoid activation for every image patch.
Following previous work \cite{Chen2018, Saito2019, Shen2019}, we attach image-level domain classifiers to different pyramid levels of the FPN to tackle the image-level domain discrepancy between different range sensor types and setups. 
By placing a GRL \cite{Ganin2015} between feature extractor and domain classifier we aim to optimize the classifier parameters to minimize the domain classification loss while simultaneously optimizing the feature extractor parameters to maximize the domain classification loss.

The instance-level domain shift can be caused by different object appearances or traffic scenarios.
We employ an instance-level domain classifier by concatenating classification and regression features as these subnets do not share weights.
Similar to the image-level, instance-level domain classifiers output domain classification maps of input patch size.
Note that there is only one instance-level domain adaptation block as the detection heads share weights across all pyramid levels.

\section{Training} \label{sec:training}
\subsection{Training Objective} \label{sec:training_objective}
Our objective
\begin{equation*}
    \min_{\boldsymbol{F},\boldsymbol{H}} \max_{\boldsymbol{D}}~\mathcal{L}_{\text{det}}(\boldsymbol{F},\boldsymbol{H}) - \lambda_1 \mathcal{L}_{\text{DA}}(\boldsymbol{F},\boldsymbol{H},\boldsymbol{D})
\end{equation*}
is to find feature extractor parameters $\boldsymbol{F}^*$ and object detection head parameters $\boldsymbol{H}^*$ that minimize the object detection $\mathcal{L}_{\text{det}}$ loss but maximize the domain adaptation loss $\mathcal{L}_{\text{DA}}$.
Conversely, we aim to find discriminator parameters $\boldsymbol{D}^*$ that minimize $\mathcal{L}_{\text{DA}}$.
Here, $\lambda_1$ is a weighting factor between both loss terms and set to $\lambda_1 = 0.5$ for all experiments.

The detection loss is only optimized by source domain training examples and ignored for target domain examples.
The domain classifier then can be seen as an adversarial discriminator that minimizes the domain classification loss while object detection model parameters maximize it.
After training, the object detection feature distributions should be aligned across domains, i.e. the feature extractor yields similar features for different domains.

The object detection loss
\begin{equation*}
    \mathcal{L}_\mathrm{det} = \frac{1}{N_\mathrm{pos}} (\mathcal{L}_\mathrm{cls} + \lambda_2 \mathcal{L}_\mathrm{box})
\end{equation*}
is a linear combination of the classification loss $\mathcal{L}_\mathrm{cls}$ implemented as focal loss \cite{Lin2017a} and a robust box regression loss $\mathcal{L}_\mathrm{box}$ implemented as smooth $\mathrm{L_1}$-loss.
Here, $N_\mathrm{pos}$ denotes the number of positive matches.
Throughout all experiments we set $\lambda_2 = 1$ and $\gamma = 2$.

We use $d_i$ as domain label for the $i$-th training example, where $d_i=0$ for a source domain example and $d_i=1$ for a target domain example.
We denote the patch-based domain classification of the $i$-th training example at the $l$-th pyramid level as $p_{il}^{(u,v)}$, where $(u,v)$ indicates the pixel of the corresponding feature map.
Then, the cross-entropy loss
\begin{align*}
    \mathcal{L}_{\text{img/ins}}^{(l)} = - \frac{1}{N H_l W_l} \sum_i \sum_{u,v} & d_i \log p_{il}^{(u,v)}          \\
                                                                                 & + \left(1-d_i\right) \left(1-\log p_{il}^{(u,v)}\right)
\end{align*}
denotes the image- or instance-level loss normalized by the number of examples $N$ and the feature dimension $H_l, W_l$ at $l$-th pyramid level.
Following \cite{Chen2018}, we also adopt a consistency regularization term that forces both levels to output similar features.
At pyramid level $l$, the consistency loss
\begin{align*}
    \mathcal{L}_{\text{cons}}^{(l)} = \frac{1}{N H_l W_l} \sum_i \sum_{u,v} \left( p_{\text{img},il}^{(u,v)} - p_{\text{ins},il}^{(u,v)} \right)^2
\end{align*}
resembles the $\mathrm{L_2}$-norm of two patches.
The final domain adaptation loss
\begin{align*}
    \mathcal{L}_{\text{DA}} = \frac{1}{L} \sum_{l=1}^L \mathcal{L}_{\text{img}}^{(l)} + \mathcal{L}_{\text{ins}}^{(l)} + \mathcal{L}_{\text{cons}}^{(l)}
\end{align*}
averages image-, instance-level and consistency loss across $L$ pyramid levels.

\subsection{Implementation Details} \label{sec:training_implementation_details}
All our models are trained in an adversarial fashion with SGD by using a gradient reversal layer \cite{Ganin2015}.
We first train a standard object detection model using only source domain data and then fine-tune it by training with target domain examples and domain adaptation loss terms.
The domain adaptive object detector is trained with an initial learning rate of $10^{-4}$ for the first 60k steps and reduced to $10^{-5}$ for the next 20k steps.
We use the momentum of 0.9 and a weight decay of $10^{-4}$.
Due to memory limitations we use a batch size of 4 where every batch consists of two training examples each from source and target domain.

\section{Evaluation} \label{sec:evaluation}
We call the process of adapting a model trained from labeled source domain examples to an unlabeled target domain as \textit{source domain to target domain}.
We evaluate our three scenarios nuScenes to KITTI, KITTI to nuScenes and KITTI to our custom data set.
During inference, our object detection models have the same number of parameters and an inference time of 52ms on a NVIDIA GeForce RTX 2080Ti.

\subsection{nuScenes to KITTI} \label{sec:from_nuscenes_to_kitti}
Out of the ten object classes in nuScenes, only the classes \textit{car}, \textit{pedestrian} and \textit{cyclist} are considered in KITTI.
Due to the difficulty of detecting extremely small objects in grid maps, we evaluate the AP for the \textit{car} class at an IoU threshold of 0.5 for the three difficulties \textit{Easy}(E), \textit{Moderate}(M) and \textit{Hard}(H).

\sisetup{table-number-alignment=center, exponent-product=\cdot, output-decimal-marker = {.}}
\begin{table}[ht]
    \centering
    \begin{tabular}{r|ccc|ccc}
                            & \multicolumn{3}{c|}{\textbf{Components}} & \multicolumn{3}{c}{\textbf{AP}}                                                        \\
                            & \textbf{Img}                             & \textbf{Ins}                    & \textbf{Cons} & \textbf{E} & \textbf{M} & \textbf{H} \\
        \hline
        Trained on nuScenes &                                          &                                 &               & 77.1       & 72.5       & 67.1       \\
        Trained on KITTI    &                                          &                                 &               & 90.0       & 87.9       & 80.1       \\
        \hline
        Exp. 1              & \checkmark                               &                                 &               & 84.7       & 72.5       & 72.1       \\
        Exp. 2              &                                          & \checkmark                      &               & 84.5       & 72.4       & 72.6       \\
        Exp. 3              & \checkmark                               & \checkmark                      &               & 85.1       & 73.9       & 73.5       \\
        Exp. 4              & \checkmark                               & \checkmark                      & \checkmark    & 86.9       & 76.5       & 76.2       \\
    \end{tabular}
    \caption{
        AP for the \textit{car} class at IoU 0.5 when trained on nuScenes labels (except \textit{Trained on KITTI}) and evaluated on KITTI.
        We denote image-level domain adaptation, instance-level domain adaptation and consistency regularization as Img, Ins and Cons, respectively.
    } \label{tab:nusc2kitti}
\end{table}
Tab.~\ref{tab:nusc2kitti} compares quantitative object detection results for the KITTI validation set for different model configurations.
We additionally provide the results of a model trained on the KITTI training set as an upper bound of AP.
Compared to the baseline, our domain-adaptive model with image- and instance-level domain-adaptation and consistency regularization yields the highest improvements, achieving an absolute performance gain of 4\% for Moderate and 9.8\% for Easy difficulty.

\begin{figure}[ht]
    \small
    \begin{scaletikzpicturetowidth}{\columnwidth}
        \begin{tikzpicture}[scale=\tikzscale]
            \begin{axis}[
                    width=\columnwidth,
                    height=0.6\columnwidth,
                    xlabel=IoU / \%,
                    ylabel=AP / \%,
                    xtick={10,20,...,90},
                    ytick={0,15,...,90},
                    grid=both,
                    cycle list name=exotic,
                    legend style={at={(0.62,0.43)},anchor=north east},
                    every axis plot/.append style={ultra thick},
                ]
                \addplot +[mark=none] plot coordinates {(20, 90.0)(30, 89.9)(40, 89.44)(50, 87.9)(60, 78.87)(70, 66.25)(80, 39.65)(90, 10.42)};
                \addplot +[mark=none] plot coordinates {(20, 85.4)(30, 85.04)(40, 83.25)(50, 76.46)(60, 63.15)(70, 28.95)(80, 5.03)(90, 0.15)};
                \addplot +[mark=none] plot coordinates {(20, 76.3)(30, 75.93)(40, 74.54)(50, 72.47)(60, 60.48)(70, 31.70)(80, 5.82)(90, 0.00)};
                \legend{Trained on KITTI\\Adapted from nuScenes\\Trained on nuScenes\\}
            \end{axis}
        \end{tikzpicture}
    \end{scaletikzpicturetowidth}
    \caption{
        AP of class \textit{car} at \textit{moderate} difficulty for our domain-adaptive detector (exp. 4) depending on the IoU threshold.
    } \label{fig:accuracy_vs_iou}
\end{figure}
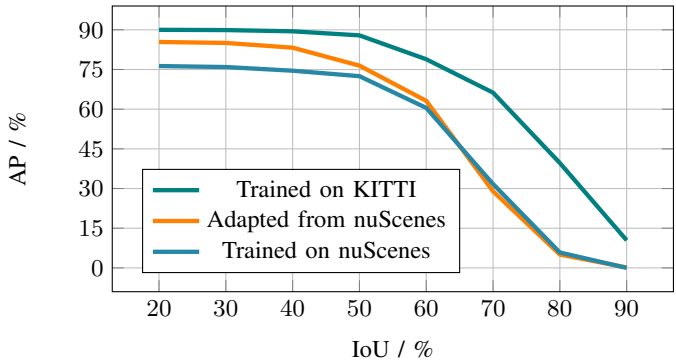
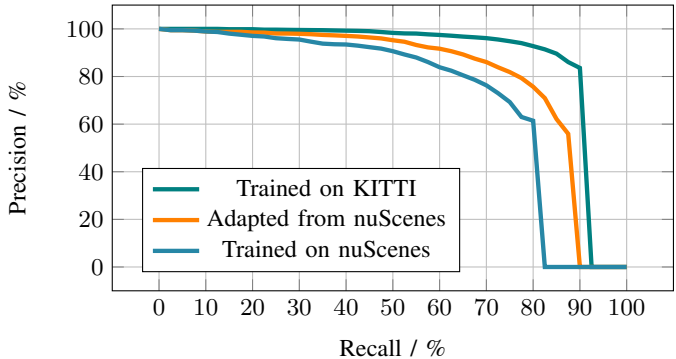
\begin{figure}[ht]
    \small
    \begin{scaletikzpicturetowidth}{\columnwidth}
        \begin{tikzpicture}[scale=\tikzscale]
            \begin{axis}[
                    width=\columnwidth,
                    height=0.6\columnwidth,
                    xlabel=Recall / \%,
                    ylabel=Precision / \%,
                    xtick={0,10,...,100},
                    ytick={0,20,...,100},
                    grid=both,
                    cycle list name=exotic,
                    legend style={at={(0.62,0.43)},anchor=north east},
                    every axis plot/.append style={ultra thick},
                ]
                \addplot +[mark=none] plot coordinates {(0, 100.0)(2.5, 100.0)(5, 100.0)(7.5, 100.0)(10, 100.0)(12.5, 100.0)(15, 99.8626)(17.5, 99.8626)(20, 99.8626)(22.5, 99.6711)(25, 99.6711)(27.5, 99.6016)(30, 99.5434)(32.5, 99.4945)(35, 99.4527)(37.5, 99.3440)(40, 99.2491)(42.5, 99.1656)(45, 99.0315)(47.5, 98.6857)(50, 98.3234)(52.5, 98.0977)(55, 98.0383)(57.5, 97.7092)(60, 97.4531)(62.5, 97.1355)(65, 96.7649)(67.5, 96.4623)(70, 96.1103)(72.5, 95.5402)(75, 94.8174)(77.5, 94.0254)(80, 92.7592)(82.5, 91.4051)(85, 89.5912)(87.5, 86.0660)(90, 83.5925)(92.5, 0)(95, 0)(97.5, 0)(100, 0)};
                \addplot +[mark=none] plot coordinates {(0, 100.0000)(2.5, 99.4536)(5, 99.4536)(7.5, 99.2727)(10, 98.9130)(12.5, 98.7578)(15, 98.7578)(17.5, 98.7578)(20, 98.5095)(22.5, 98.4356)(25, 98.0583)(27.5, 98.0392)(30, 97.9335)(32.5, 97.7649)(35, 97.4713)(37.5, 97.2877)(40, 97.0628)(42.5, 96.8045)(45, 96.4054)(47.5, 95.9978)(50, 95.2331)(52.5, 94.5959)(55, 93.2369)(57.5, 92.2330)(60, 91.6772)(62.5, 90.6262)(65, 89.2709)(67.5, 87.5178)(70, 86.0670)(72.5, 83.8638)(75, 81.8619)(77.5, 79.3521)(80, 75.7095)(82.5, 70.9319)(85, 62.1355)(87.5, 55.9292)(90, 0.0000)(92.5, 0.0000)(95, 0.0000)(97.5, 0.0000)(100, 0.0000)};
                \addplot +[mark=none] plot coordinates {(0, 100.0000)(2.5, 99.4536)(5, 99.4536)(7.5, 99.2727)(10, 98.9130)(12.5, 98.6957)(15, 98.0216)(17.5, 97.5460)(20, 97.0628)(22.5, 96.8047)(25, 96.0888)(27.5, 95.7854)(30, 95.5302)(32.5, 94.6314)(35, 93.8053)(37.5, 93.5484)(40, 93.4447)(42.5, 92.9603)(45, 92.3251)(47.5, 91.7153)(50, 90.6281)(52.5, 89.2423)(55, 87.9842)(57.5, 86.0791)(60, 83.8846)(62.5, 82.4383)(65, 80.5111)(67.5, 78.6286)(70, 76.2890)(72.5, 72.9917)(75, 69.2757)(77.5, 63.0060)(80, 61.4283)(82.5, 00.0000)(85, 00.0000)(87.5, 00.0000)(90, 00.0000)(92.5, 00.0000)(95, 00.0000)(97.5, 00.0000)(100, 00.0000)};
                \legend{Trained on KITTI\\Adapted from nuScenes\\Trained on nuScenes\\}
            \end{axis}
        \end{tikzpicture}
    \end{scaletikzpicturetowidth}
    \caption{
        Precision-recall curve for \textit{car} detections on the KITTI validation set at an IoU threshold of 0.5 and \textit{moderate} difficulty.
    } \label{fig:precision_recall}
\end{figure}
Overall, there is a performance gap between our best model and the model trained on KITTI which is depicted for different IoUs in Fig.~\ref{fig:accuracy_vs_iou}.
We observe that the AP of our domain-adaptive model decreases towards the baseline model AP for increasing IoU thresholds.
This indicates that the precision of box regression does not benefit as much from domain-adaptation.
However, domain adaptation increases the recall compared to our baseline which is depicted in Fig.~\ref{fig:precision_recall}.

\begin{figure*}[ht]
    \centering
    \begin{subfigure}[b]{0.31\textwidth}
        \includegraphics[width=\textwidth]{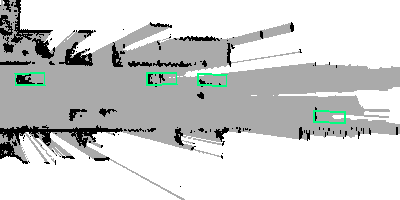}
        \caption{Without domain adaptation}
        \label{fig:eva_kitti_before}
    \end{subfigure}
    \hspace{10pt}
    \begin{subfigure}[b]{0.31\textwidth}
        \includegraphics[width=\textwidth]{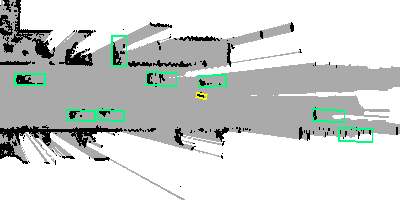}
        \caption{With domain adaptation (exp. 4)}
        \label{fig:eva_kitti_after}
    \end{subfigure}
    \hspace{10pt}
    \begin{subfigure}[b]{0.31\textwidth}
        \includegraphics[width=\textwidth]{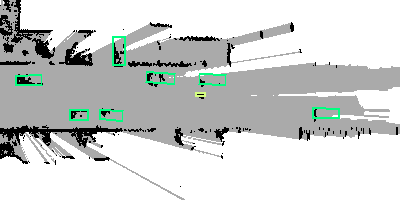}
        \caption{Ground truth}
        \label{fig:eva_kitti_gt}
    \end{subfigure}
    \caption{
        Qualitative object detection results on the same KITTI validation scenario without and with domain adaptation (exp. 4).
        Cars, cyclists and pedestrians are depicted in green, yellow-green and yellow, respectively.
    } \label{fig:nusc2kitti}
\end{figure*}

Fig.~\ref{fig:nusc2kitti} depicts qualitative object detection results with and without domain adaptation compared to ground truth.
We observe that car can be detected in the adapted model which are missed in the baseline.

\subsection{KITTI to nuScenes} \label{sec:from_kitti_to_nuscenes}
In our experiments, we only consider the mAP, ATE and ASE (see Sec.~\ref{sec:related_work_datasets}) for evaluation.
We do not report the Average Orientation Error (AOE) because our box encoding does not distinguish between forward and backward orientation, whereas nuScenes considers the orientation between $0$ and $2\pi$.

Tab.~\ref{tab:kitti2nusc} depicts the quantitative object detection results.
We observe that the image-level domain adaptation alone yields better results than only instance-level domain adaptation.
As for the experiments nuScenes to KITTI, we observe the best results for all components enabled.
However, we observe a high absolute performance gap of 22.4\% between our domain adaptive object detector and the model trained on nuScenes.
The reason might be the high data set variability, i.e. unexpected scenarios, which cannot be considered as domain gap.
Also, there is no significant improvement in ATE and ASE compared to the baseline.

\sisetup{table-number-alignment=center, exponent-product=\cdot, output-decimal-marker = {.}}
\begin{table}[ht]
    \centering
    \begin{tabular}{r|ccc|c|cc}
                            & \multicolumn{3}{c|}{\textbf{Components}} & \textbf{mAP} & \multicolumn{2}{c}{\textbf{TP-Metrics}}                                      \\
                            & \textbf{Img}                             & \textbf{Ins} & \textbf{Cons}                           &      & \textbf{ATE} & \textbf{ASE} \\
        \hline
        Trained on KITTI    &                                          &              &                                         & 8.6  & 0.27         & 0.5          \\
        Trained on nuScenes &                                          &              &                                         & 36.9 & 0.20         & 0.46         \\
        \hline
        Exp. 5              & \checkmark                               &              &                                         & 11.4 & 0.27         & 0.50         \\
        Exp. 6              &                                          & \checkmark   &                                         & 9.4  & 0.28         & 0.50         \\
        Exp. 7              & \checkmark                               & \checkmark   &                                         & 11.0 & 0.28         & 0.49         \\
        Exp. 8              & \checkmark                               & \checkmark   & \checkmark                              & 14.5 & 0.29         & 0.49         \\
    \end{tabular}
    \caption{
        Quantitative results for the \textit{car} class when trained with labeled KITTI data (except \textit{Trained on nuScenes}) and evaluated on the nuScenes validation set.
    } \label{tab:kitti2nusc}
\end{table}

\subsection{KITTI to Custom Setup} \label{sec:custom_setup}
The custom data set is collected by our experimental vehicle which carries four Velodyne VLP-16 at different mounting positions.
The grid maps are generated by fusing range measurements within a time interval of 50ms and then transformed into one common vehicle frame.
This leads to slightly misaligned observations, depending on the velocity difference between vehicle and other traffic participants or static environment.
These misalignments usually deteriorate the object detection performance, especially the recall of small objects.
Additionally, the performance is negatively influenced by the different sensor setup.

\begin{table*}[ht]
    \centering
    \begin{tabular}{cccc}
         & \textbf{Scenario 1}                                                                            & \textbf{Scenario 2} & \textbf{Scenario 3} \\
        \toprule
        \rotatebox[origin=c]{90}{Before}
         & \raisebox{-.5\height}{\includegraphics[width=0.28\textwidth]{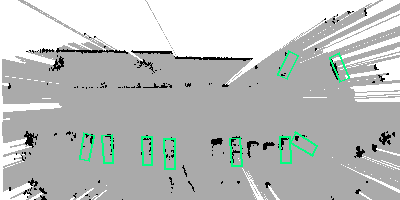}}
         & \raisebox{-.5\height}{\includegraphics[width=0.28\textwidth]{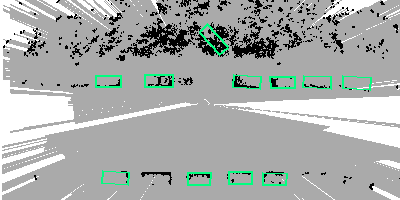}}
         & \raisebox{-.5\height}{\includegraphics[width=0.28\textwidth]{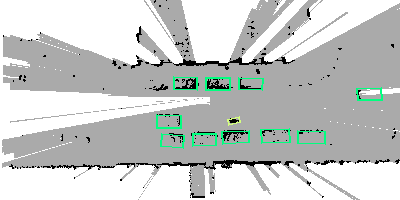}}                                             \\
        \midrule
        {\rotatebox[origin=c]{90}{After}}
         & \raisebox{-.5\height}{\includegraphics[width=0.28\textwidth]{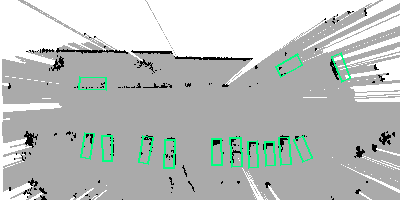}}
         & \raisebox{-.5\height}{\includegraphics[width=0.28\textwidth]{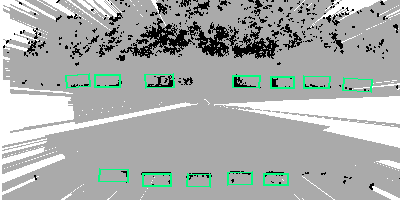}}
         & \raisebox{-.5\height}{\includegraphics[width=0.28\textwidth]{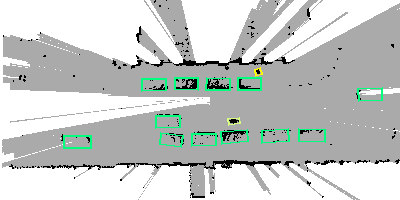}}
    \end{tabular}
    \caption{
        Qualitative results before and after domain adaptation on three scenarios from a custom data set.
        Cars, cyclists and pedestrians are depicted in green, yellow-green and yellow, respectively.
        We observe that the adapted model yields a higher recall and a slightly higher box regression accuracy.
    } \label{tab:eva_bertha}
\end{table*}

Tab.~\ref{tab:eva_bertha} compares object detectors trained on KITTI with and without custom sensor setup adaptation.
We observe a much higher recall and a slight improvement in box regression accuracy when the object detector is trained with all domain adaptation components.

\section{Conclusion} \label{sec:conclusion}

We presented our approach to adapt single shot FPN object detectors for grid maps to a custom sensor setup using image- and instance-level domain adaptation.
Our quantitative experiments with the nuScenes and KITTI benchmarks show an increased recall and a slight improvement in precision for our domain-adaptive object detector.

In the future, we hope to increase model generalization by extending our approach to one target and multiple source domains.
Also, to reduce the additionally needed model capacity when trained on two domains, we would like to use separate translation units for each domain.
During model deployment we can then pick only the relevant translation unit and save memory and computation time.

\bibliographystyle{IEEEtran}
\bibliography{root}
\end{document}